# The State of the Art when using GPUs in Devising Image Generation Methods Using Deep Learning

## Yasuko Kawahata[1]

[1] Department of Mathematical Informatics Graduate School of Information Science and Technology, The University of Tokyo 7-3-1 Hongo, Bunkyo-ku, Tokyo 113-8656, Japan.

**Abstract**

Deep learning is a technique for machine learning using multi-layer neural networks. It has been used for image synthesis and image recognition, but in recent years, it has also been used for various social detection and social labeling. In this analysis, we compared (1) the number of Iterations per minute between the GPU and CPU when using the VGG model and the NIN model, and (2) the number of Iterations per minute by the number of pixels when using the VGG model, using an image with 128 pixels. When the number of pixels was 64 or 128, the processing time was almost the same when using the GPU, but when the number of pixels was changed to 256, the number of iterations per minute decreased and the processing time increased by about three times. In this case study, since the number of pixels becomes core dumping when the number of pixels is 512 or more, we can consider that we should consider improvement in the vector calculation part. If we aim to achieve 8K highly saturated computer graphics using neural networks, we will need to consider an environment that allows computation even when the size of the image becomes even more highly saturated and massive, and parallel computation when performing image recognition and tuning.

## 1. About Deep Learning(2015~2016)

Deep learning is a technique for machine learning using multilayer neural networks. In the past, technologies such as fuzzy networks and neural networks have been used for image synthesis [1] and image recognition [2] that incorporate elements of opportunity learning, but in recent years, they have quickly become one of the technologies that have attracted attention. Due to the progress of learning algorithms and the availability of a large amount of input data, the performance of neural networks has improved dramatically compared to conventional neural networks. Many fields where the market is expected to expand in the future, such as automatic driving, IoT (Internet of Things), and robotics, have a close relationship with machine learning or artificial intelligence, and are expected to develop into future applications.

The general recognition seems to be caused by the acquisition of DeepMind by Google, and the appearance of Google, IBM, and Caffe. In the 2014s, simple character recognition and image recognition packages such as Theano and Pylearn2 appeared, but after 2015, more powerful image recognition and processing packages using CUDA from NIVIDIA started to appear. From Google, APIs such as Cloud Vision API [3], Tensoflow, MXnet, and H2O, as well as packages and libraries that can be implemented more easily using Python and R have also appeared. Chainer, which uses the same Python library as Tensorflow, has also been introduced. Unlike Caffe, Chainer and Tensorflow are relatively easy to integrate with CUDA libraries, but they are even faster because they can be integrated with Python's acceleration package Cyton. It is also highly versatile in that it can be used with cuDNN, a GPU library for deep learning developed by NIVIDIA. METAMIND has started to use it in the field of x-ray image diagnosis in medicine[4]. Furthermore, in Japan, UEI announced the DEEPstation DK-1, a personal computer dedicated to deep learning, and Alpaca has started distributing APIs for image recognition and tools for predicting financial data[5, 6]. In January 2016, Microsoft also distributed toolkits such as CNTK, and we can say that it is becoming more familiar. As for Microsoft, in a project called "Catapult," we have been conducting research on improving computational efficiency by attaching FPGAs to servers in data centers. In the research of Catapult, they are not only building an accelerator engine by attaching FPGAs to the CPUs of servers, but also challenging the structure of a two-dimensional torus network using FPGAs to handle the communication between nodes. In a report in February 2016, we succeeded in accelerating the image recognition flow (FPGA) in Google's Tensorflow by a factor of although the performance of FPGA is usually 1/5 of that of GPU, it can be compensated by scaling up. In deep learning, a CNN (Convolutional Neural Network), which mimics a neural system, is used, and a huge amount of computation is required to evaluate and train the input. We aim to improve the performance by one order of magnitude or more, with a cost increase of less than 30% and a power increase of less than 10%, by using the abundant resources of FPGAs to perform these calculations. Microsoft is aiming to popularize deep learning by providing the FPGA accelerator as a software library for users to use in the future[7]. In the field of deep learning, it can be said that there is a lot of discussion on how to speed up the learning of a large number of images, how to compete with other models for more detailed image recognition, and how to improve the development environment.

## 2. Expression techniques using Deep Learning(2015~2016)

As described in the previous section, the idea of using multilayer neural networks such as neural networks itself has been around for a long time, and these techniques have been used not only in industrial applications but also in the field of artistic expression[9]. With the advocacy of Genda et al. (1980), a new technique for expressing graphics has emerged, in which mathematical models are devised based on graphics such as images to be generated, using phenomena and things that occur in society and nature as input [9-13]. In addition, representation techniques that generate new images by evolutionary computation reflecting neural network operations in real time have been used in situations such as the self-organizing CG generation by Kawaguchi (2001)[14]. In addition, the environment on GPUs is becoming more and more important in the generation of graphics that require large-scale operations. Since the dawn of computer graphics, Genda et al(1980). have proposed a number of techniques for computer animation, such as the production of animation using random scan displays that take advantage of machine power[13]. And new graphic expressions based on the application of deep learning techniques are expected to increase in the future[15]. GPU-based operations are also necessary in the field of image recognition, which has outstanding results in deep learning compared to conventional machine learning. Nowadays, optimized GPU libraries have appeared to speed up image recognition techniques for deep learning such as cuDNN [16]. In this report, we compare the results of CPU and GPU benchmarks to see the difference in graphics generation between CPU and GPU benchmarks, in order to explore the potential of deep learning in image recognition, image generation, and graphics representation. By conducting this test, we would like to consider the scale, data size and machine for graphic representation using deep learning and batch plotting of analysis results of large-scale data in the future.

## 3. Consider which package to Use Cases(2015~2016)

As described in Section 1, there are many libraries, packages, and toolkits related to deep learning, but in this case, we used Chainer, which has been reported to be an effective benchmark when using GPUs among the packages available until 2015[17]. Under the environment of a 6-core Intel Core i7-5930K CPU @ 3.50GHz + NVIDIA Titan X + Ubuntu 14.04 x86_64, the following results were obtained, and we decided to use Chainer in this report.

| Model | Alexnet[18] | Overfeat[19] | OxfordNet[20] | Googlenet V1[21] |
|---|---|---|---|---|
| Chainer | 177(sec) | 620(sec) | 885(sec) | 687(sec) |
| TensorFlow | 277(sec) | 843(sec) | 1510(sec) | 1084(sec) |
| Caffe (native) | 324(sec) | 823(sec) | 1068(sec) | 1935(sec) |

Table 1: Execution times (in seconds) for various deep learning models under 6-core Intel Core i7-5930K CPU @ 3.50GHz + NVIDIA Titan X + Ubuntu 14.04 x86_64.

)[18-21].

## 4. About Chainer(2015~2016)

Chainer is a library for implementing neural networks developed by Preferred Networks [17]. This package is characterized by its support for CUDA. It is also possible to implement various types of neural networks, such as convolutional and recurrent, using Python.

## 5. Points to consider when installing Chainer(2015~2016)

When you install Chainer, you need to install some related libraries.
You also need to install CUDA and cuDNN. Note that you need to register an account for cuDNN at the CUDA official website, answer a questionnaire about your usage, and download the cuDNN for your environment [16]. In addition, since Chainer depends on the Python library, it is also important to have an environment where the pip command can be used as follows.

```
$ su
$ yum install python-setuptools
$ wget http://peak.telecommunity.com/dist/ez_setup.py
$ /usr/local/bin/python ez_setup.py
$ easy_install pip
$ easy_install numpy
$ easy_install six
$ easy_install Mako
$ easy_install scipy
$ pip install scikit-learn
lspci | grep -i nvidia
```

Table 2: Codes for configuration (excerpt)

Before using GPU and cuDNN in Chainer, you should check your graphics board with the above command. Also, make sure to run Devicequery before running the program. The following assumes that CUDA7.5 is installed. Note that we recommend you to install Chainer 1.5 or later. The reason is that Chainer1.5 or later has no dependency with Pycuda, and thus there is no difficulty in passing through cuDNN in the later versions.

CPATH=$CPATH:/usr/local/cuda-7.5/include PATH=$PATH:/usr/local/cuda-7.5/bin CUDA_ROOT=/usr/local/cuda-7.5 export LD_LIBRARY_PATH=/usr/local/cuda-7.5:$LD_LIBRARY_PATH which nvcc ./deviceQuery sudo sh -c " cd /root/NVIDIA_CUDA-7.5_Samples/1_Utilities/deviceQuery/; ./deviceQuery " CPATH=$CPATH:/usr/local/cuda-7.5/lib64 PATH=$PATH:/usr/local/cuda-7.5/lib64 CUDA_ROOT=/usr/local/cuda-7.5/lib64 export LD_LIBRARY_PATH=/usr/local/cuda-7.5/lib64:$LD_LIBRARY_PATH CPATH=$CPATH:/usr/local/cuda-7.5/bin PATH=$PATH:/usr/local/cuda-7.5/bin CUDA_ROOT=/usr/local/cuda-7.5/bin export LD_LIBRARY_PATH=/usr/local/cuda-7.5/bin:$LD_LIBRARY_PATH CPATH=$CPATH:/usr/local/cuda-7.5/include PATH=$PATH:/usr/local/cuda-7.5/include CUDA_ROOT=/usr/local/cuda-7.5/include export LD_LIBRARY_PATH=/usr/local/cuda-7.5/include:$LD_LIBRARY_PATH

Table 3: Codes for setting up the system (excerpt)

## 6. Models Handled(2015~2016)

In this report, we compare the computational results of the NIN model and the VGG model [22] based on A Neural Algorithm of Artistic Stlye proposed as a method for image generation in deep learning, and the benchmarks of CPU, GPU, and cuDNN. The basis of our model is the Convolutional Learning Model. The basis of our model is to generate an image using Convolutional Neural Network.

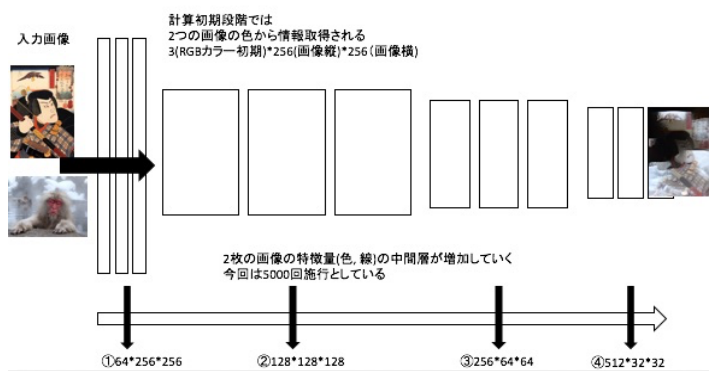

Figure 1: Picture of the model flow.

The numbers in Figure 1 mean [number of channels * height * width]. The basic model is based on a Convolutional Neural Network, and is generated from parameters that are intermediate between the main features of the two images.

Assume that the input images are two images of 128*128 pixels, and the number of channels is 3 since the input images are RGB tri-color. As the layers increase, the number of channels increases, and the intermediate values of line thickness and line thinness are acquired and redrawn for representation. In this figure, we used 256*256 as an example, but we found that the algorithm works even if the resolution is changed, and the processing time becomes longer as the size is larger. In this algorithm, the output from the intermediate layers 1-4 is used.

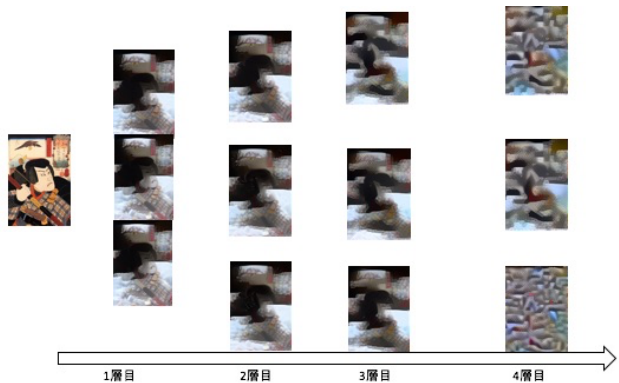

Figure 2: Channel flow

The following are the generated images.

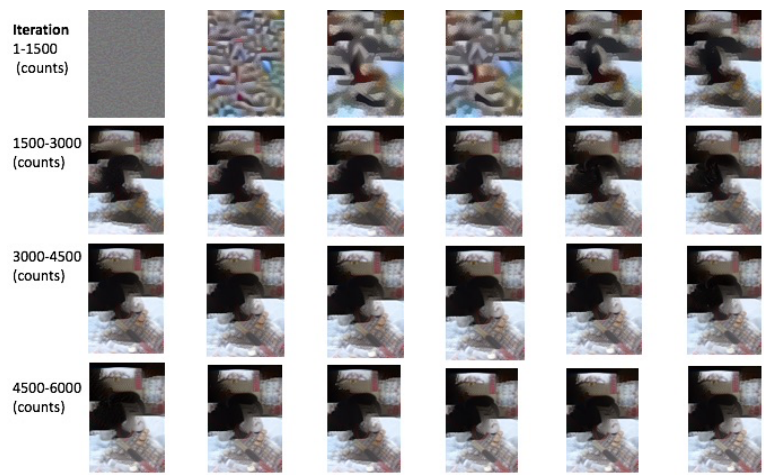

Figure 3: Generated image (VGG model)

In addition, the correlation between the three channels output by Stylenet is calculated, and the features of the entire image, as well as the thickness and fineness of the lines, are represented in each intermediate layer.

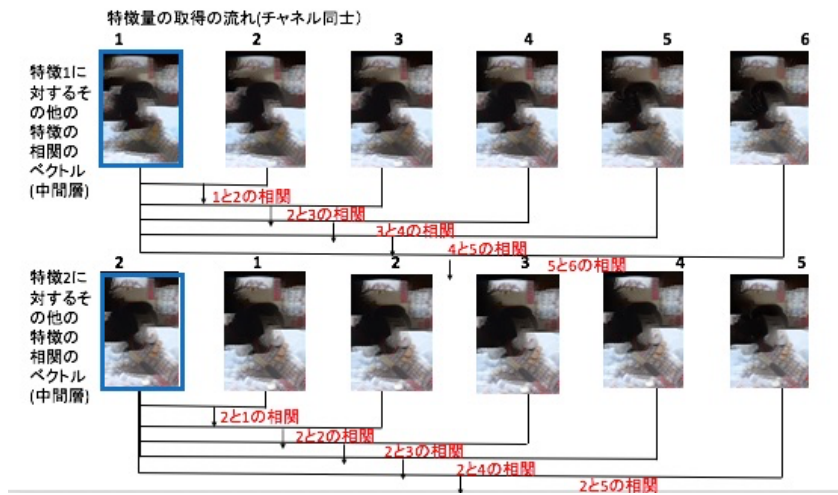

Figure 4: Flow of feature acquisition.

By outputting these results, the computational results are reflected in the generated image. The idea of the overall objective function is to correlate the difference between the image to be synthesized and the image of the middle layer to be minimized, and the difference in color and line between the Stylenet and the image, and to make a vector.

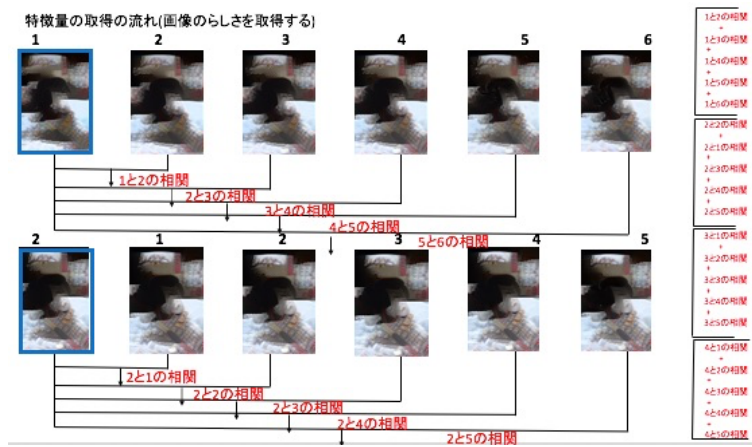

Figure 5: The flow of obtaining image features.

In this way, the difference between the former and the latter style images can be measured in both shallow and deep layers, and information such as fine brush strokes can be extracted in the shallow layers, while larger spatial patterns can be extracted in the deep layers.

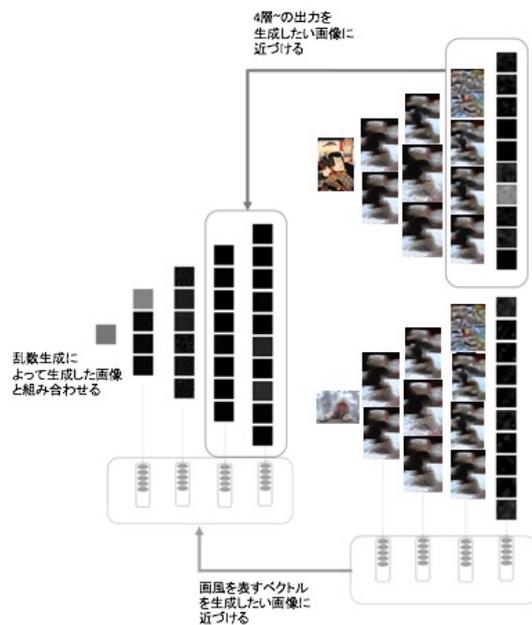

Figure 6: The overall flow

As shown in Fig. 6, in the initial stage, the image was generated by gradually repeating the calculation, starting from a figure reflecting only RGB information, gradually reflecting the special quantity of lines, and obtaining the intermediate quantity of line depth.

## 7. Calculation Results

The environment used in this report is Intel(R) Core(TM) i5-4460 CPU @ 3.20GHz + NVIDIA Corporation GM206 [GeForce GTX 960] + CentOS Linux release 7.2.1511 (Core). In this section, we compare (1) the number of Iterations per minute between the GPU and the CPU when the VGG model is used and when the NIN model is used, and (2) the number of Iterations per minute by the number of pixels when the VGG model is used, using an image with 128 pixels.
Through (1) and (2), we discussed the number of calculations per minute on the GPU according to each model and the number of pixels handled.

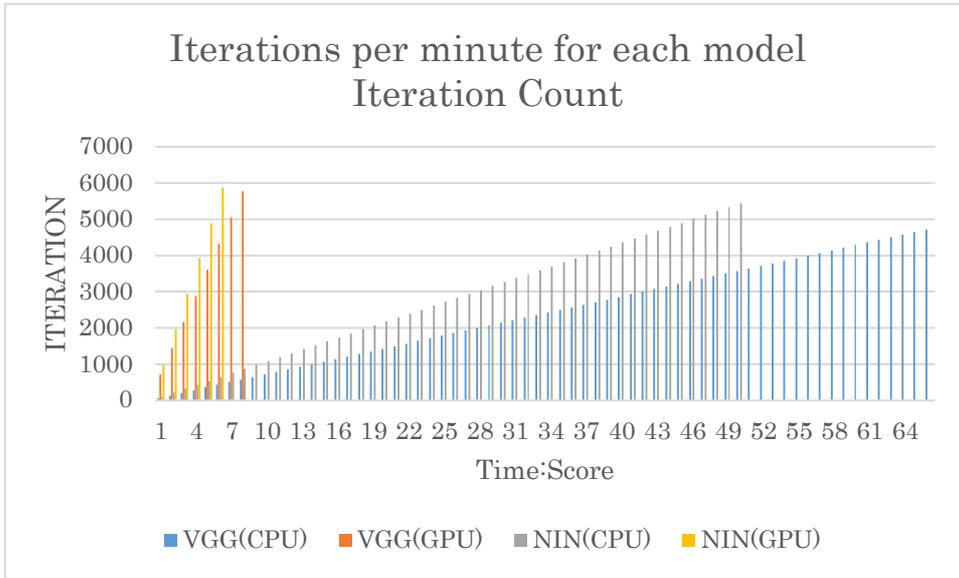

Figure 7: Comparison of GPU and CPU when using VGG model and NIN model. Number of iterations per minute, using an image with 128 pixels.

| Time(minutes) | VGG(CPU)/ITERATION(Per Count) | VGG(GPU) /ITERATION (Per Count) | NIN(CPU) /ITERATION (Per Count) | NIN(GPU) /ITERATION (Per Count) |
|---|---|---|---|---|
| 1(Per Minutes) | 71.4 | 721.9 | 108.9 | 980 |
| 2 | 142.9 | 1443.7 | 217.8 | 1960 |
| 3 | 214.3 | 2165.6 | 326.7 | 2940 |
| 4 | 285.7 | 2887.4 | 435.6 | 3920 |
| 5 | 357.1 | 3609.3 | 544.4 | 4900 |

Table 4: Comparison of the number of Iterations per minute between the GPU and CPU when using the VGG model and when using the NIN model (Iterations up to 5 minutes, using an image with 128 pixels).

Using an image with 128 pixels, we compared the number of iterations per minute between the GPU and CPU when using the VGG model and when using the NIN model. In Figure 7 and Table 4, the comparison of VGG and NIN models by model and environment is omitted because the cuDNN processing is supposed to improve the speed, but it is faster when only GPU mode is used.

As can be seen from Figs. 2 and 3, VGG is a model that strictly reflects the accuracy of the generated image and the intermediate amount of pixels, so the computation was somewhat slower than NIN, with fewer iterations per minute even in GPU mode. The issue here is how to speed up the computation of VGG, which will be a theme in the future.

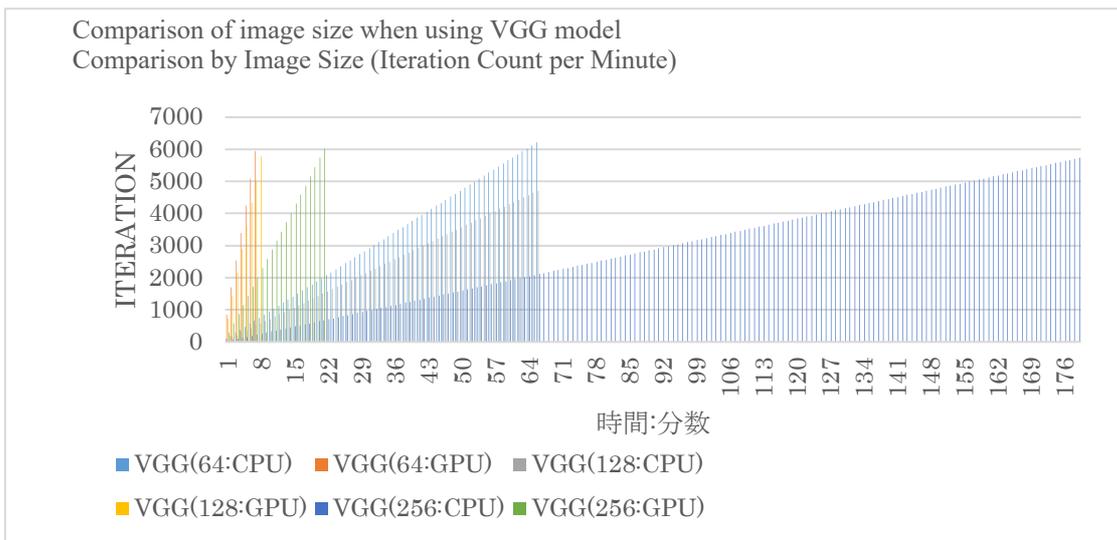

Figure 8: Comparison by image size when using VGG model.

(Number of Iterations per minute)

| Time (minutes) | ITERATION times of VGG (CPU in case of 64 pixels) | TERATION times of VGG (GPU in case of 64 pixels) | ITERATION times of VGG (CPU in case of 128 pixels) | ITERATION times of VGG (GPU for 128 pixels) | ITERATION times of VGG (CPU in case of 256 2-pixel number) | ITERATION of VGG (GPU for 256 pixels) |
|---|---|---|---|---|---|---|
| 1 | 94.2 | 847.8 | 71.4 | 721.9 | 32.0 | 286.7 |
| 2 | 188.3 | 1695.5 | 142.9 | 1443.7 | 64.1 | 573.3 |
| 3 | 282.6 | 2543.3 | 214.3 | 2165.6 | 96.1 | 860.1 |
| 4 | 376.8 | 3391.0 | 285. 7 | 2887.4 | 128.1 | 1146.7 |
| 5 | 471.0 | 4238.8 | 357.1 | 3609.3 | 160.1 | 1433.4 |

Table 5: Comparison of Iterations per minute by pixel count when using VGG model (Iterations up to 5 minutes)

We also compared the number of Iterations per minute by the number of pixels when using the VGG model. In the above comparison of models and environments in the VGG model in Fig. 8 and Table 5, we omitted the cuDNN processing in this example because it should have improved the speed, but it was faster in the GPU mode only. This is an issue to be addressed in the future.

However, when the number of pixels was changed to 256, the number of Iteations per minute was reduced, and the processing time was increased by a factor of about three. As a future prospect, when we consider the case of generating highly saturated images with the size of 1920 pixels or more, we can consider that we should consider improvements in the vector calculation part, because in this case, the core dumping occurs when the number of pixels is 512 or more.

## 8. Prospect

In this report, we were able to generate images with clearer material detection by using GPU operations, but the problem is that the deep learning flow itself requires a large amount of computation time, and the image size of 512 pixels causes core dumping in the environment of this report. In the future, if we aim to achieve 8K highly saturated computer graphics using neural networks such as those of Genda(1990) and Kawaguchi(2000), it will be necessary to consider the construction of an environment in which computation is possible even when the size of the image becomes even more saturated and massive, and parallel computation when performing image recognition and tuning.In the future, we can expect to use GPUs for machine learning of big data using deep learning, and we would like to discuss image generation based on image learning in this report, as well as application examples.

## Acknowledgments

This research is the Result of Research Project 15J06703(Japan Society for the Promotion of Science PD) "Mathematical empirical analysis of past complex social phenomena using historical materials". In addition, the Graduate School of Information Science and Engineering, the University of Tokyo, who provided the GPU analysis environment for this research. I would like to thank Professor NAKATA Toshiyuki of the Social ICT Research Center.